# Variation and Synthetic Speech

Corey Miller, Orhan Karaali and Noel Massey

## 1. Introduction

As high-quality synthetic speech becomes more and more of a reality, the question of how to introduce what we have learned about variation becomes a question of critical importance. If indeed variation is as important to understanding and explaining language as we have been saying it is, how should we apply our knowledge to speech generation systems? In order to approach this issue, we need to adjust our perspective to describing variation in such a way that it is possible to allow it to serve as a model for a synthetic voice.

After motivating the investigation of variation with respect to the development of speech synthesis systems, we review past treatments of variation in such systems. We then present the approach to variation taken by the Motorola text-to-speech system, highlighting the pronunciation lexicon, phonematizer (or letter-to-sound converter) and the postlexical module. The sensitivity of the synthesizer to the nature of individual variation results in an adaptable system capable of reproducing the natural voice qualities and structured variation of different speakers.

### 1.1. Treatment of variation in speech synthesis

Modern text-to-speech systems utilize a pronunciation dictionary for pronunciation of most words (Liberman and Church 1992). Thus, the kinds of pronunciations provided in such a dictionary are critical to the synthesizer's performance. Several studies have aimed to capture dialect variation in synthetic speech by providing a means for storing different pronunciations for different dialects in the synthesizer's lexicon (Williams and Isard 1997, Fitt 1995, 1997, Bladon et al. 1987). In contrast to these phonological approaches, Hertz and Huffman (1992) show how detailed information on the phonetic implementation in each dialect is required.

## 2. Why model segmental variation?

As Labov and his colleagues in the Cross Dialectal Comprehension project have shown, segmental differences between American dialects can often result in serious misunderstandings (Labov 1989). Labov showed that Chicagoans, for exam-ple, had a marked local advantage in interpreting speech from Chicago, as compared to people from Birmingham and Philadelphia. Even when discourse context might have been thought to be a useful disambiguator, listeners often chose nonsense words to describe phonetic forms alien to their dialects (e.g. *budgeroom* over *bedroom* for [bʌdruwm] ).

Given that synthetic speech has been shown to be less understandable than natural speech (Pisoni 1997), it would seem that avoiding miscomprehension due to dialect differences is an important goal for synthetic speech. Therefore, if a synthesizer is to be used in a particular community, it may be a good idea to employ a synthesis of the local dialect, thus insuring maximum intelligibility.

Nearness to the dialect of the listener may also be thought likely to engender greater acceptability for synthetic speech. For example, Williams and Isard (1997) suggest that a Scottish bank telephone service with synthetic speech would be better off having a Scottish accent than an English one. However, all local dialects are not necessarily prized by their speakers. Labov has shown in New York (1966) and elsewhere that locals react negatively to phonological characteristics of their local dialects.

It is important to distinguish between lexical and postlexical variation in the studies we have been citing, particularly with respect to intelligibility and comprehension. Labov's studies of vowel shifts (Labov 1991), many of which played a significant part in his cross-dialectal comprehension studies, are of the lexical variety – that is, words in a speaker's lexicon would presumably be stored with the vowels in question. In contrast, much sociolinguistic research has concentrated on postlexical variants, such as studies of *t,d* (e.g. Guy 1980) deletion in English or *s* deletion in Spanish (e.g. Poplack 1980).

Another dimension along which variation can be categorized is that of inter- and intra-speaker variation. The approach to modeling individual postlexical variation that we will present should be contrasted with approaches in both sociolinguistics and speech technology that model inter- rather than intraspeaker variation. In sociolinguistics, Guy (1980) showed that in the case of *t,d* deletion, variation in the individual tends to mirror that of the speech community. Howevr, Van de Velde and van Hout (1997) show how different explanations of the variation in Dutch *n* deletion result from an examination of individual, as opposed to aggregate, data. Approaches to pronunciation modeling in speech recognition (e.g. Tajchman et al. 1995), attempt to predict a range of dialect and postlexical phenomena across speakers.

### 2.1. Attempts to add variation to synthetic speech

We will now review two studies in synthetic speech which have sought to employ models of postlexical variation. Portele (1997) presented five phoneticians and six naive listeners with six sentences produced by a German speech synthesizer. Each synthetic sentence was produced in two forms, one with reduced vowel variants and the other without reduced vowel variants. The phoneticians preferred the reduced forms, while the naive listeners preferred the unreduced forms. According to Portele, "Synthetische Sprache wird jedoch anders perzipiert als natürliche Sprache, bei letzterer werden kanonische Realisierungen kaum toleriert, während synthetische Sprache durchaus 'deutlich' klingen darf"[1]. From this study it appears that language professionals may have different, perhaps paternalistic, notions about what people want to hear from a speech synthesizer, and these judgments may not necessarily accord with those of the general public.

It is also possible that synthetic speech is still too unrealistic (i.e. not human sounding enough) to support very human characteristics like vowel reduction. That is, people may not be comfortable with a machine behaving in a manner they deem too human. Even a machine like Data on *Star Trek: The Next Generation*, who is human in so many respects, evinces his "machineness" by his failure to use contractions.

Sorin (1991) reports on the problem of mute *e* in French speech synthesis. She reports that the problem had previously been "hidden" due to the poor quality of synthetic speech. In other words, when a synthesizer's basic segmental intelligibility is still questionable, it probably does not make sense for researchers to spend time worrying about issues of postlexical variation. Now that quality had improved, she sought to address variation in mute *e,* in an effort to make the CNET synthesizer sound more natural.

Most of the time, in colloquial French, the mute *e* is elided if its presence does not facilitate the pronunciation of the word or word sequence. A previous version of the CNET synthesizer only elided those mute *e*'s that appear at the end of polysyllabic words. While it was thought that intelligibility was maximized by this procedure, listeners perceived the behavior as dysfluent or stumbling.

---

[1] "Apparently synthetic speech is perceived differently from natural speech. In natural speech, canonical realizations are hardly tolerated, while synthetic speech must sound thoroughly precise."

Sorin ran an experiment with conditions with differing treatments of mute *e* by the synthesizer. The main generalizations derived from this test were:

(1) When mute *e* is elided in synthetic speech, correct word identification is substantially lower than that obtained with natural speech pronounced the same way.
(2) Systematic pronunciation of intermediate mute *e* in synthetic speech leads to identification scores virtually identical to those obtained with natural speech, in which all examples of mute *e* are elided.

However, when subjects were asked to give a preference rating to the synthetic versions with and without elision, the one with all pronounced mute *e*'s pronounced was judged worst – it did not sound fully natural. While Sorin's results show that introduction of some natural postlexical variation can lead to improved user preference, they also show that there may be an intelligibility cost to the introduction of that variation.

### 2.2. Acceptability and intelligibility divergence

Nusbaum et al. 1995b report a similar result to Sorin's in an evaluation of several speech synthesizers, including two from Motorola. Nusbaum et al. performed two sets of experiments, one set designed to determine the ranking of synthesizers with respect to intelligibility and another set designed to determine the ranking of the synthesizers with respect to acceptability.

For the acceptability experiment, subjects were played speech in several domains where synthetic speech might be used, such as finding out movie times, or getting account information from a bank. Subjects were asked to rate the acceptability of each sentence for that domain on a scale of 1 to 7. A rating of 1 was described as "I would not use a service that used this voice", while a rating of 7 was described as "I would prefer to use a service with this voice". A rating of 4 was described as "This voice is acceptable". For the intelligibility experiment, subjects were played single words from several synthesizers and one human talker. They were instructed to type into a computer the word they heard in normal spelling, and to type a nonsense word if it did not sound like a word.

The Motorola synthesizers did reliably better (at $p < .05$) in acceptability than the other synthesizers. In contrast, the Motorola synthesizers performed reliably worse than two of the other synthesizers in intelligibility. In the past, acceptability had been mostly a measure of intelligibility, since in general synthesizers suffered from severe intelligibility problems. Today, intelligibility differences between synthesizers are less glaring, so acceptability judgments may be based more on judgments of naturalness, voice quality and prosody. The fact that the Motorola synthesizers were lower ranked for intelligibility than acceptability is cited as proof that listeners were judging acceptability and intelligibility along different lines.

Nusbaum et al. were surprised that the most acceptable synthesizer was not found to be the most intelligible. They report that this is the first such discrepancy that they are aware of. However, it seems to us that Sorin experienced the same result – the more natural sounding synthesis was in fact not the most intelligible.

There is a growing body of research on so-called "clear speech", which has properties such as lack of reduction and slow tempo that make it generally more intelligible than normal speech (e.g. Picheny et al. 1985, 1986). Clear speech, which is used for example when speaking with foreigners or the hard of hearing, does not sound natural. Therefore, it provides another example of a possible asymmetry between naturalness and intelligibility. However, when comprehension of longer passages is concerned, it is our contention that more natural speech should draw attention away from *how* things are being said to *what* is being said, therefore yielding better comprehension (cf. Ralston et al. 1995). We intend to pusue this hypothesis in future work.

## 3. The Motorola Speech Synthesis System

Speech synthesis techniques that attempt to model the speech of an individual, such that that individual's voice could be recognizable from the synthetic speech have become common. The Motorola speech synthesizer (Karaali et al. 1997, Corrigan et al. 1997) is among these, as it uses neural networks trained on the acoustic and durational properties of a recorded speech database of an individual to generate natural, intelligible speech recognizable as from that individual. If another voice is desired, all that is necessary is to train the neural networks on a recorded speech database of that voice.

### 3.1. Run-time operation

As shown in Figure 1, the Motorola speech synthesizer contains a Text Processing unit which builds a textual representation of a text that is designated by a user to be spoken aloud. This text might consist of e-mail messages, web pages or the results of a database query. That textual representation contains information about word parts of speech and syntactic phrase boundary information. The pronunciation of each of the words identified by the text processor's tokenizer is looked up in the Lexical Database. In cases of homographs like *live* /lɪv/ and *live* /lajv/, part of speech or semantic information is used to determine the appropriate lexical entry to use.

**Figure 1**

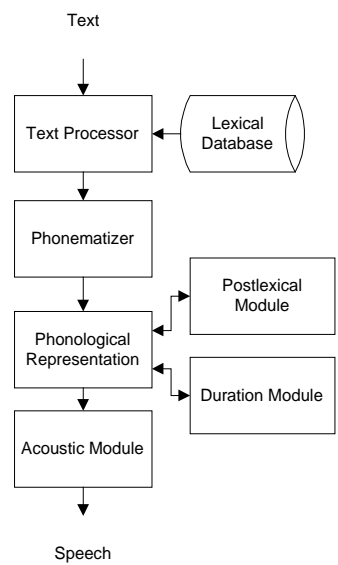

Of course, some words that are to be spoken will not be present in our lexicon, or any lexicon, no matter how large. These "out-of-dictionary" words may include some business, personal, and place names, neologisms and misspellings. The textual representation is passed to a phonematizer, to be described below, which assigns pronunciations to such out-of-dictionary words.

At this point, a preliminary phonological representation of the utterance is built from the textual representation. The phonological representation organizes the utterance into a hierarchical prosodic phonological structure, including phones, syllables and phonological words, as well as higher order constituents. The phonological representation is then submitted to a

Postlexical Module where lexical pronunciations derived from the lexicon are converted to postlexical pronunciations typical of the speaker whose voice is being modeled. The modifications that take place in the postlexical module at present involve only segmental insertions, deletions and substitutions, such as *t,d* deletion.

The phonological representation, having been modified to reflect postlexical phonology, is then submitted to a Duration Module (Corrigan et al. 1997) which assigns durations to each phone, according to the speech of individual being modeled. Finally, the phonological representation with constituent durations is submitted to an Acoustic Module (Karaali et al. 1997), which transforms the phonological representation into spectral parameters which are synthesized into a speech waveform.

### 3.2. Training overview

**Figure 2**

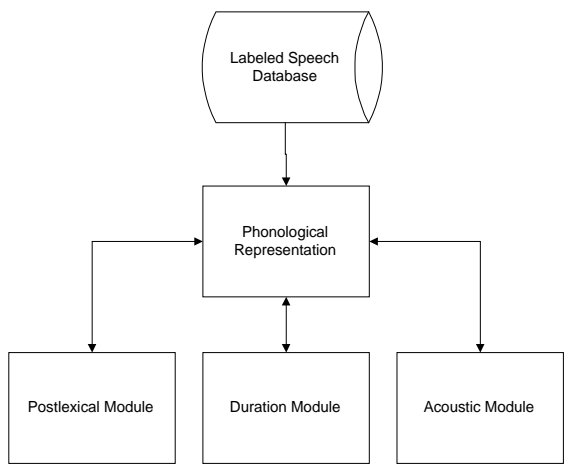

The Phonematizer, Postlexical Module, Duration Module and Acoustic Module all use neural networks trained on relevant data. This data-oriented approach facilitates customization of the modules for individual voices. As shown in Figure 2, the Postlexical Module, Duration Module and Acoustic Module all train on Phonological Representations similar to the ones upon which they are tested in run-time operation. Rather than being derived from text, as in the case of run-time operation, the phonological representations are derived from a labeled speech corpus from one individual. In fact, it is the task of the system's developers to attempt to have the phonological representations derived from text equate as much as possible with the phonological representations derived from the labeled speech corpus.

The speech corpora are labeled segmentally in a manner similar to the TIMIT database (Seneff and Zue 1988). Each utterance is also labeled prosodically according to the ToBI guidelines (Beckman and Elam 1997). Finally, stress and syllable information as well as syntactic boundaries, word type (function or content) and word prominence (related to part of speech, cf. O'Shaughnessy 1976) information are all provided by hand.

Each of the three neural networks trained on phonological representations derived from the labeled speech corpus take full advantage of the rich contextual information available, as well as the boundary information. This will be described in detail here with respect to the postlexical module.

In contrast to the postlexical, duration and acoustic modules which are retrained on each new voice, the phonematizer is only trained once per language variety. For example, all varieties of American English can use the same lexical database and phonematizer trained on it. The postlexical module takes care of customizing the lexical entries to the dialect spoken by the speaker.

### 3.3. Phonematizer

### 3.3.1. Lexical database

The phonematizer is a neural network trained on a pronunciation lexicon, which is a relational lexical database created from three source lexica: The *Carnegie Mellon Pronouncing Dictionary* (Weide 1995), *Moby Pronunciator II* (Ward 1996) and *COMLEX English pronouncing lexicon* (also known as Pronlex, Linguistic Data Consortium 1995)[2]. The dictionaries we used tended to list variant pronunciations associated with particular orthographies with no annotation, regardless of whether the variants were sociolinguistic (*e.g.* /kɔt/ and /kɑt/ for *caught*), based on part of speech distinctions (such as *live* /lɪv/ and *live* /lajv/) or semantics (such as

---

[2] The creation of this database owes much to William Thompson, Peter Viechnicki and Erica Zeinfeld.

*lead* /lɛd/ and *lead* /lijd/). This kind of dictionary is apparently adequate for speech recognition, where an acoustic input needs to be mapped to the proper orthography.

For speech synthesis, it is important to pronounce input orthographies correctly. Synthesizers have modules which can tag incoming words for part of speech (Church 1988) or semantics (Yarowsky 1997) in order to help select the right pronunciations. Of course, this requires that lexica tag non-homophonous homographs for part of speech or semantics, if required. The purpose of combining the three source lexica was to identify and tag for part of speech or semantics as many non-homophonous homographs as possible. Transcription protocols are being developed, based on McLemore (1995) and Schmidt et al. (1993), in order to make dictionary entries stemming from different source lexica consistent.

In addition, sociolinguistic variants were removed in favor of one fairly consistent dialect.[3] The idea behind removing sociolinguistic variation from the lexicon was to have a lexicon representing one plausible dialect of American English, and one from which various dialects and styles could be derived. One way to ensure that the base lexicon could be transformed into various dialects is to choose pronunciations reflecting a dialect that has distinctions rather than mergers or neutralizations for such phenomena as vowels before /r/ (*e.g. marry, Mary, merry*) or voiced and voiceless "wh" as in *which* and *witch*. Of course, a lexicon based on these principles will end up reflecting no real dialect in particular. Bladon et al. (1987) and Williams and Isard (1997) both describe lexical representations from which both British and American dialects could be derived. Williams and Isard (1997) base their approach on Wells' (1982) lexical sets.

We refer to the level of transcription found in the dictionary as the "lexical" level. We contrast this with the "postlexical" level in accordance with lexical phonology (Kiparsky 1982, 1985).[4] Lexical pronunciations are characterized by their appropriateness for use in isolation, in contrast to postlexical pronunciations which are appropriate in connected speech. Lexical pronunciations show a lack of possible vowel reduction, and they do not feature segmental reflexes of postlexical rules such as flapping.

---

[3]This is particularly challenging and has not yet been completed. For example, it is difficult to decide on the appropriate distribution of /ɑ/ and /ɔ/ to store in the lexicon.

[4] Our terminological distinction between *lexical* and *postlexical* is in contrast, for example, to Tajchman et al. (1995), who refer to the kinds of transcriptions we are calling lexical as postlexical because they contain separate, complete entries for inflections and derivations of root forms. Given the relative simplicity of English morphology (compared to languages like Finnish or Spanish), many English natural language systems do without morphological analysis and morphophonological rules, relying instead on exhaustive lexical listing of forms.

### 3.3.2. Phonematizer Training

The purpose of the phonematizer is to create lexical pronunciations for words that are not found in the dictionary. There are traditions in English (Carney 1997), speech technology (Elovitz et al. 1976) and psycholinguistics (Seidenberg 1989) discussing the correspondence between English orthography and pronunciation. Our approach is inspired by work using neural networks trained on dictionaries to determine pronunciations from orthography (e.g. Sejnowski and Rosenberg 1987). For the experiments described below, we will be using Monnet, a neural network simulator being developed at Motorola (Karaali et al. 1996).

One of the most important aspects of working with neural networks is using an appropriate coding scheme for the input and output representations. This is an area where domain knowledge, in this case linguistic, is critical. The more useful contextual information that is provided the network, the better the chance is that it will perform robustly when asked to generalize beyond the training data.

The first step in training the neural network to learn letter-phone correspondences is to align letters and phones in a meaningful way. In order to do this, we used a dynamic programming alignment algorithm (Kruskal 1983). While this algorithm is described for aligning sequences from the same alphabets, we needed to define a specialized cost function reflecting the likelihood that particular letters correspond with particular phones.

In addition to providing the network with aligned pronunciations and orthographies for each word in the dictionary, we provided the network with feature information for both phones and letters. We defined features for a letter to be the union of the features of the phones that that letter might represent. So the features for the letter 'c' would be the union of the features for /s/ and /k/.

While we are now achieving competitive results with the phonematizer, we believe that improved results will come from simplifying the phonological representations found in the dictionary, for example by removing allophones. We are confident that such simplifications will not detract from the quality of our ultimate output, due to the postlexical module, to be described next.

**3.4. Postlexical Module**

As discussed above, our dictionary and letter-to-sound procedures convert from orthographies to lexical pronunciations. Lexical pronunciations, such as those found in the dictionary sources described above, are not suitable for use in a speech synthesizer that aims to sound both natural and intelligible. For example, stringing together such lexical pronunciations can sound stilted and overprecise. We will describe a procedure for generating postlexical pronunciations from lexical pronunciations, such as those in our lexical database, or those generated by the phonematizer that was trained on them, that results in more natural, and we believe, more comprehensible speech.

There are several different perspectives from which to consider postlexical variation. One way is to look at the structural symbolic, or transcription, differences between lexical and postlexical representations. For example, certain kinds of deletion can be viewed as the deletion of a symbol, while flapping could be viewed as the substitution of one symbol for another. Of course, such a symbolic approach might be faulted for ignoring the finer grained featural changes that are taking place. Some postlexical variation might best be considered from an acoustic perspective.[5] We believe that the combination of the postlexical module and the acoustic module described previously represent a reasonable compromise between the discrete and gradient aspects of postlexical phonology.

The manner in which the speech database that we are using was labeled will have an effect on the way in which we view postlexical variation. Lander (1995) points out how both the TIMIT and the more recent CSLU (Center for Spoken Language Understanding, Oregon Graduate Institute) labeling systems for English are largely phonemic, but include several allophones, which are deemed "spectrally distinct" and "frequently occurring". For example, voiceless stops are indicated with the same symbol, regardless of whether they are aspirated, but flaps, fronted /u/ and two reduced vowel qualities can be indicated.

In the case of the subphonemic vowel symbols, we intend to compare the neural network's performance with and without such labels, in an effort to see how much the network's learning of postlexical variation is assisted by explicit labeling. We will attempt to determine the relative value of achieving postlexical variation by symbolic means versus acoustic, phonetic means. For example, we will examine the formant values of fronted /u/ when given a special label, versus letting the neural network determine phonetic allophony on the basis of a single *phonemic* label.

---

[5] See Liberman and Pierrehumbert (1984) and Kiparsky (1985) for more discussion along these lines.

### 3.4.1. Postlexical Module Training

The database of postlexical forms used in the experiments below consists of over 700 sentences of different types spoken by a 38 year old male speaker who has lived in both Florida and Chicago. The sentences were divided into training and testing subsets of 3550 and 405 words respectively. The speech was labeled at a fairly narrow phonetic level in a manner similar to the TIMIT database (Seneff and Zue, 1988).

We aligned the speech database with lexical pronunciations from the lexicon used in phonematizer training. This alignment used the same dynamic programming algorithm as described above in the context of the phonematizer's letter-phoneme alignment. A principal difference was in the substitution cost function employed. As described above, when sequences from different alphabets are to be aligned, it is important to explicitly define the cost of substituting particular members of those alphabets. In the case of the lexical-postlexical conversion, the alphabets are largely the same, but not identical. For example, phones like [ɾ] and [ʔ] appear in the postlexical alphabet, but not in the lexical alphabet.

The neural network input coding employed a window of nine phones. The use of a window allows critical contextual information to be accounted for by the net. For example, Pierrehumbert and Frisch (1997), in a study on synthesizing allophonic glottalization, point out how a running window on prosodic information is critical for determining when to invoke that particular postlexical phenomenon. For each phone, we provided feature information for both lexical and postlexical phones. In addition, for each phone, distance to word, phrase, clause and sentence boundaries was included. In future work, we intend to make use of the prosodic hierarchy, provided by the database's ToBI annotation, in the neural network encoding.

### 3.4.2. Postlexicalizer results

In the unseen testing subset of the data, the lexical and postlexical phones were identical 70% of the time. That is, 30% of the time, there was a different postlexical phone from lexical phone. Testing of a neural network trained as described above on a segment of the database that was excluded from training resulted in 87% correct prediction of postlexical phones. This indicates that while substantial learning took place, there is still room for increased learning.

Table 1 compares the realizations of lexical /d/ in both the labeled corpus and the postlexical neural network trained on it, while Table 2 examines lexical /t/. While the proportions predicted by the network appear to be reasonable approximations of the actual data, an exhaustive performance analysis of the network, taking phonological context into account, remains for future work.

**Table 1: Realizations of /d/**

| Realization of /d/ | Number of occurrences in labeled corpus | Number of occurrences predicted by postlexical neural network |
|---|---|---|
| closure + release | 51 (40%) | 62 (49%) |
| closure only | 14 (11%) | 19 (15%) |
| release only | 12 (10%) | 15 (12%) |
| flap | 12 (10%) | 10 (8%) |
| deleted | 14 (11%) | 20 (16%) |

**Table 2: Realizations of /t/**

| Realization of /t/ | Number of occurrences in labeled corpus | Number of occurrences predicted by postlexical neural network |
|---|---|---|
| closure + release | 68 (54%) | 69 (55%) |
| closure only | 26 (21%) | 28 (22%) |
| release only | 12 (10%) | 8 (6%) |
| flap | 4 (3%) | 3 (2%) |
| glottal stop | 8 (6%) | 8 (6%) |
| deleted | 7 (6%) | 8 (6%) |

### 3.4.3. Postlexical phenomena learned

Table 3 illustrates several of the postlexical phenomena learned by the postlexical neural network. These phenomena are all well-known postlexical phenomena of English. Many of them occur variably in the speech of individual speakers along a stylistic continuum.

**Table 3: Postlexical phenomena learned by network**

| Phenomenon | Lexical | Postlexical | Orthography |
|---|---|---|---|
| unreleased stops | fɛd | fɛd̚ | fed |
| glottalized vowels | ænd | ʔænd | and |
| glottalized consonants | stret | streʔ | straight |
| *d* deletion | ænd fɑlo | æn fɑlo | and follow |
| *t* deletion | əbrʌpt stɑrt | əbrʌp̚ stɑrt̚ | abrupt start |
| destressing and assimilation | ði tæŋk | ðɨ tæŋk | the tank |
| destressing and assimilation | ði waʲndɪŋ | ðə waʲndɪŋ | the winding |
| *t* flapping | dɚˑti | dɚˑɾi | dirty |
| nasal flapping | kɔrnɚˑ | kɔrɾ̃ɚˑ | corner |
| *h* voicing | ɪn hɚˑ | ɪn ɦɚˑ | in her |
| schwa epenthesis | kɚˑlz | kɚˑlz | curls |

### 3.4.4. Dialect/Labeling/Lexical Idiosyncrasies learned

Table 4 illustrates several phenomena learned by the postlexical network that would probably not be considered postlexical in a description of natural language. These phenomena are the results of the neural network learning some of the idiosyncrasies of the particular lexicon-labeler-speaker triad at hand. For example, neutralization of vowels before /r/ is probably not a postlexical process in the speaker's actual grammar, since his mental lexicon probably stores neutralized versions of these vowels. However, when coupled with a general American English pronunciation lexicon, as described above, it is necessary to perform this neutralization actively, which is what the postlexical network does.

Table 4: Dialect/labeling/lexical idiosyncrasies learned by network

| Phenomenon | Lexical | Postlexical | Orthography |
|---|---|---|---|
| marry/merry/Mary merger | bærəl | bɛɹl̩ | barrel |
| schwa deletion/metathesis? | čıldrənz | čıldərnz | children's |
| laxing before /r/ | kornɚ | kɔrɾɚ | corner |
| vocalic consonants | pudəlz | pʉdl̩z | poodles |
| wh voicing | ʍaʲlst | waʲlst | whilst |
| /u/ fronting | dun | dʉn | dune |

## 4. Conclusion

We have shown that recognizing the structure of variation in a voice to be synthesized is an important element in correctly modeling that voice. We have described the ascription of linguistic levels in the Motorola Speech Synthesizer, where lexical representations are reserved for the lexicon and the phonematizer that is trained on it, and postlexical representations are reserved for the postlexical module. This organization allows the lexicon and phonematizer to serve for most instantiations of a rather broad language variety, and a postlexicalizer to be automatically retrained for each individual voice. A neural network based postlexicalizer was described which achieved 87% correct postlexical phone assignment on unseen data. We believe that the accuracy of individual variation that this architecture permits for speech synthesis will result in increased naturalness and acceptability of our synthesizer.

Speech Processing Research Laboratory
Motorola Inc.
1301 E. Algonquin Road
Schaumburg, IL  60196
coreym@ccrl.mot.com
karaali@ccrl.mot.com
massey@ccrl.mot.com